\documentclass[runningheads]{llncs}
\usepackage[T1]{fontenc}
\usepackage{graphicx}
\usepackage{booktabs}
\usepackage[misc]{ifsym}
\newcommand{\corr}{(\Letter)}
\usepackage{mwe}
\usepackage{microtype}
\usepackage{subcaption}
\usepackage{booktabs} 
\usepackage{amsmath}
\usepackage{longtable} 
\usepackage{bm}
\usepackage{xcolor}
\usepackage{amsfonts}
\usepackage{hyperref}
\newcommand{\light}[1]{\textcolor{gray!70}{#1}}

\begin{document}

\title{Learning by Shifting: \\ Temporal View Construction for Time Series Contrastive Learning}

\titlerunning{Learning by Shifting}

\author{Abdul-Kazeem Shamba\inst{1} \corr \and
Kerstin Bach\inst{1}  \and
Gavin Taylor\inst{2}}

\authorrunning{A.K. Shamba et al.}

\institute{Norwegian University of Science and Technology, Norway \email{\{abdul.k.shamba,kerstin.bach\}@ntnu.no}
\and
United States Naval Academy, USA \\
\email{taylor@usna.edu}}

\maketitle              

\begin{abstract}
Supervised learning demands large quantities of labeled data, a bottleneck that is expensive and reliant on domain-specific expertise. Self-supervised learning, particularly contrastive learning, has emerged as a compelling alternative, enabling rich representation learning directly from unlabeled data. Yet its success hinges critically on the design of positive and negative sample pairs. Existing approaches for time series rely on hand-crafted augmentations and masking heuristics that embed strong domain assumptions, often limiting generalization across diverse temporal patterns and potentially introducing spurious correlations. In this work, we challenge this paradigm by demonstrating that explicitly encoding temporal shift invariance through a simple, deterministic view construction is sufficient to learn strong representations for time series classification. By exploiting temporal structure, our method, \textit{Shift Invariant Feature Training} (ShiFT), achieves state-of-the-art performance on six diverse real-world time series benchmark datasets, as well as the UCR and UEA archives, while reducing training time. Beyond empirical performance, we present a systematic analysis of contrastive learning dynamics in time series settings, examining the effects of batch size and the number of negatives on downstream performance. Our findings provide practical insights for designing efficient contrastive learning frameworks for time series representation learning. The source code is publicly available at \href{https://github.com/sfi-norwai/ShiFT}{https://github.com/sfi-norwai/ShiFT}.

\end{abstract}

\begin{figure}[h!]
\includegraphics[width=\textwidth]{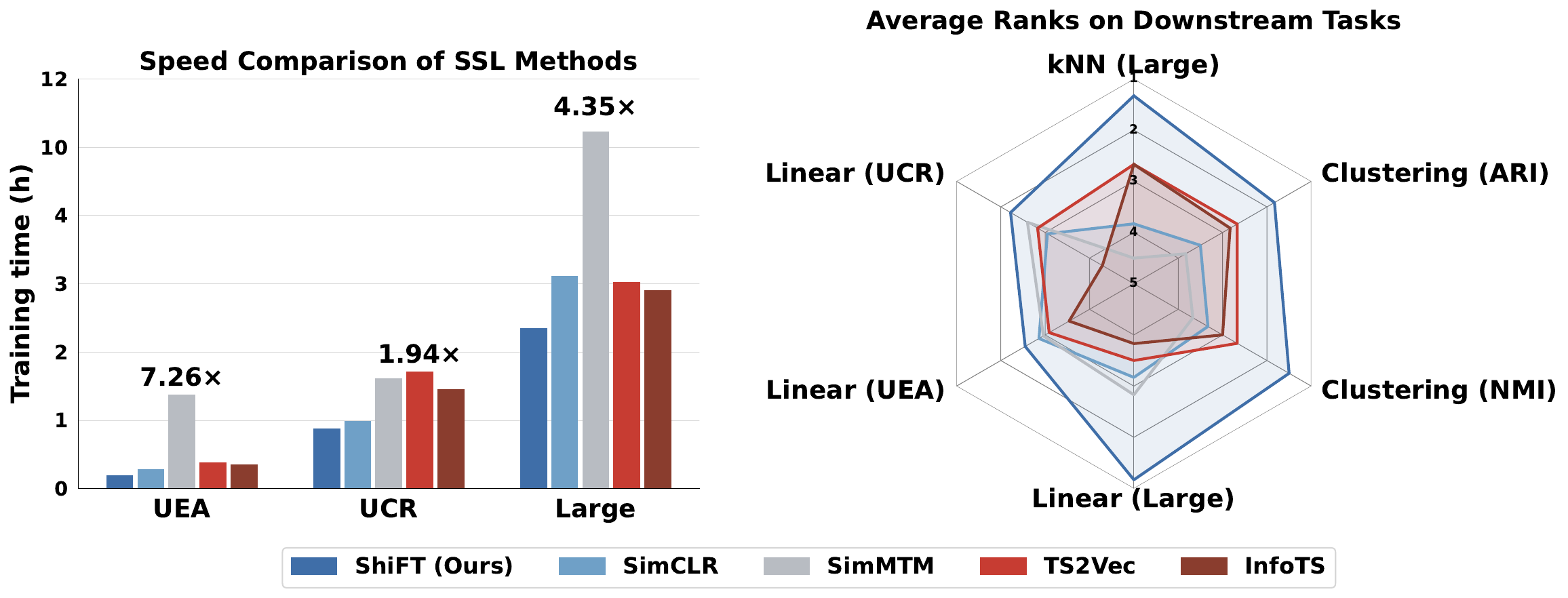}
\caption{Training efficiency and downstream performance of SSL methods. \textit{Left:} Total training time (hours) across three benchmark settings -- six large-scale datasets (${>}20\text{k}$ samples), the UCR archive, and the UEA archive. Speedup multipliers are reported relative to ShiFT. \textit{Right:} Average rank across six downstream evaluation tasks, where lower rank indicates better performance. ShiFT achieves the fastest training time across all three settings while maintaining the best average rank on all downstream tasks, demonstrating that simplicity in view construction does not compromise representational quality.} 
\label{fig:radar}
\end{figure}

\section{Introduction}

Time series data is generated continuously across a variety of critical domains, such as physiological monitoring in healthcare, motion capture in robotics, and vibration sensing in industrial systems. Despite the abundance of such data, meaningful labels remain scarce as annotation is costly, time-consuming, and demands expertise rarely available at scale, constraining the supervised approaches that dominate time series analysis today. This core bottleneck has driven growing interest in self-supervised learning (SSL), which learns rich representations directly from unlabeled data, enabling strong downstream performance without heavy reliance on manual annotation. 

Early SSL approaches relied on proxy objectives, commonly referred to as pretext tasks, such as solving jigsaw puzzles, reconstructing masked segments, or predicting applied transformations. While such tasks encourage feature learning, they introduce an inherent misalignment: the model optimizes for an objective structurally disconnected from any downstream task, risking representations that are well-adapted to the pretext signal but transfer poorly in practice. Contrastive learning (CL) addresses this limitation more directly. Rather than solving a fixed proxy task, contrastive methods learn by attracting semantically similar views (positive pairs) while repelling dissimilar ones (negative pairs) in the embedding space. This formulation allows the desired invariances to be encoded directly into the learning objective, bypassing the pretext-downstream mismatch. In computer vision (CV), this has proven highly effective: augmentation-based view generation produces semantically consistent positive pairs at scale, enabling representations that generalize broadly across downstream tasks \cite{simclr,moco,dino}.

However, extending these gains to time series poses distinct challenges. Unlike natural images, time series lack a canonical augmentation space: perturbations such as jittering, scaling, or time warping implicitly assume which signal properties are semantically irrelevant, assumptions that rarely hold uniformly across domains. Beyond augmentation, methods that rely on complex masking hierarchies \cite{simmtm,ts2vec} introduce additional computational overhead and domain-specific inductive biases, limiting cross-domain transferability. Choosing proper contrastive methodologies for a given domain is a complex and essential engineering choice that often determines the usefulness of the representation. These design choices couple representation quality to engineering effort, undermining the scalability that makes SSL appealing in the first place.

We argue that this complexity is not warranted. In many time series tasks, the class label of a sequence depends not on when a pattern occurs, but on whether it occurs, implying a natural invariance to temporal translation. Two temporally shifted segments of the same signal may capture the same underlying phenomenon, and a well-trained encoder should reflect this. Rather than approximating this invariance through stochastic augmentation, we propose \textit{Shift Invariant Feature Training} (ShiFT), which encodes this invariance directly and deterministically into the view construction process. Specifically, ShiFT generates positive pairs by splitting sequences into temporally shifted windows with a fixed overlap. This simple strategy preserves the signal's structural properties while encouraging the model to learn representations that are invariant to temporal translations. Our main contributions are summarized as follows:
\begin{itemize}
    \item We show that a simple, deterministic shift-based view selection strategy can rival or outperform complex methods that rely on hand-crafted augmentations or masking schemes for time series classification, while reducing training time.
    
    \item We provide empirical insights into how linear and non-linear projection heads, batch size, and the number of negatives affect representation quality in time series contrastive learning.

    \item We validate ShiFT on six large-scale time series datasets (${{>}}20\text{k}$ samples), referred to as Large, as well as the UCR and UEA archives, demonstrating strong and consistent performance across diverse temporal domains.
    
\end{itemize}

\section{Related Work}
\textbf{Self-Supervised Learning.} Self-supervised learning has emerged as a powerful paradigm for learning representations from unlabeled data by defining pretext tasks that exploit the inherent structure of the data itself. Early approaches formulated these as proxy objectives such as predicting image rotations \cite{rotation}, solving jigsaw puzzles \cite{jigsaw}, or reconstructing masked regions \cite{inpainting}, but suffer from an inherent misalignment between the pretext objective and the downstream task, risking representations that generalize poorly. Contrastive learning addresses this limitation by replacing fixed proxy tasks with a flexible objective that directly encodes desired invariances into the representation space. SimCLR \cite{simclr} demonstrated that strong data augmentation combined with a normalized temperature-scaled cross-entropy loss is sufficient to learn highly transferable representations. MoCo \cite{moco} introduced a momentum encoder and memory bank to decouple the batch size from the number of negatives, improving scalability. DINO \cite{dino} extended this to a self-distillation setting without explicit negatives, using a teacher-student architecture to prevent representation collapse. Collectively, these works established that representation quality is highly sensitive to view construction, a finding that directly motivates our work.

\textbf{Contrastive Learning in Time Series.} Motivated by these advances, several works have adapted contrastive learning to the time series setting, with varying assumptions about temporal structure and semantic similarity. TS2Vec \cite{ts2vec} introduces a hierarchical contrastive framework operating across instance and temporal dimensions using randomly masked overlapping windows. InfoTS \cite{infots} adaptively selects augmentations that maximize mutual information between views via a meta-learning approach. SimMTM \cite{simmtm} departs entirely from the contrastive paradigm, framing SSL as a masked-reconstruction task over manifold-aware patch aggregations. While these methods achieve strong performance, they share a common limitation: reliance on hand-crafted augmentations and complex sampling strategies introduces domain-specific assumptions and computational overhead, limiting scalability and cross-domain transferability.

\textbf{View Construction for Contrastive Learning.} The design of positive pairs is central to contrastive learning, as it defines what invariances the encoder is required to learn. In CV, views are generated through compositions of stochastic augmentations and the choice of augmentation pipeline has been shown to critically determine representation quality \cite{simclr}. In the time series domain, this is less principled: temporal signals are sensitive to transformation type and magnitude, and augmentations borrowed from vision can distort the structure that distinguishes classes. Several works design domain-specific augmentations \cite{infots,dou2026autodatimeseries,ts2vec}, but this specialization limits generalizability. Deterministic view construction, where positive pairs are defined by a fixed structural relationship rather than stochastic perturbation, avoids augmentation bias. Our work demonstrates that a simple deterministic shifted-split strategy encodes sufficient inductive bias for discriminative temporal representation learning across diverse domains, without any domain-specific assumptions.

\section{ShiFT Method}

\subsection{Problem Definition}

Let $\mathcal{X} = \{x_1, x_2, \ldots, x_N\}$ denote a batch of $N$ unlabeled time series instances, where each $x_i \in \mathbb{R}^{T \times C}$, with $T$ denoting the sequence length and $C$ the number of channels. Our objective is to learn an encoder $f_\theta : \mathbb{R}^{T \times C} \rightarrow \mathbb{R}^d$ that maps each instance to a $d$-dimensional representation $z_i = f_\theta(x_i)$, such that the learned embeddings are discriminative and transferable to downstream classification tasks without any labeled supervision.

We adopt the contrastive learning framework, in which two views $\mathcal{X}^{(1)}$ and $\mathcal{X}^{(2)}$ are constructed from each instance, and the encoder is trained to align the representations of views derived from the same instance while contrasting those from different instances. The key design choice, and the central focus of this work, is the view construction strategy.

\begin{figure}[t]
\includegraphics[width=\textwidth]{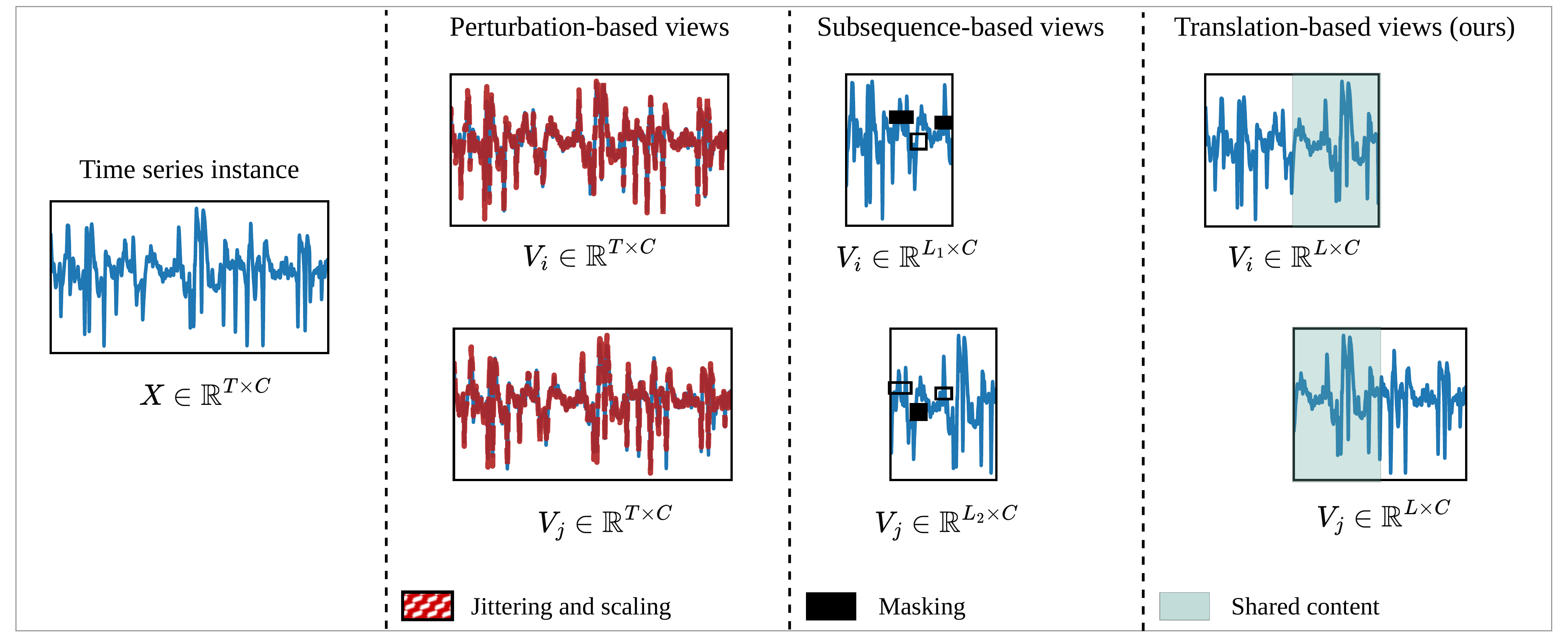}
\caption{Comparison of view construction strategies for contrastive learning 
in time series. Perturbation-based views (left) apply jittering and scaling 
to the full sequence, introducing synthetic distortion. Subsequence-based 
views (centre) extract variable-length windows via masking, discarding 
observed signal. Translation-based views (right, ours) construct two 
overlapping shifted splits of equal length $L$, preserving the original 
signal while sharing a common region of semantic content (shaded).}
\label{fig: perturb}
\end{figure}

\subsection{What Makes a Good View in Time Series Contrastive Learning}

The view construction strategy is the most consequential design decision in contrastive learning, as it implicitly defines the invariances the encoder is required to learn. A view $V(\cdot)$ is a stochastic or deterministic transformation applied to an input instance. For contrastive learning to support downstream classification, a view must satisfy three competing constraints.

\paragraph{Label Preservation.} A view must not alter the semantic class of the original instance: $y(x) = y(V(x)).$ If the transformation can change the class label, the contrastive objective injects label noise by design, directly degrading downstream performance.

\paragraph{Class Informativeness.} The view must retain sufficient discriminative structure: $I(V(x);\, y) \text{ is high,}$ where $I(\cdot\,;\,\cdot)$ denotes mutual information. Invariance alone is insufficient, a view that discards class-relevant variation will produce embeddings that are well-aligned but not discriminative.

\paragraph{Non-Triviality.} The two views must be sufficiently distinct:
\begin{equation}
    V_1(x) \not\approx V_2(x), \quad I(V_1(x);\, V_2(x)) \text{ is high}.
\end{equation}
If the views are too similar, the contrastive objective degenerates into identity matching, and the encoder learns no useful structure.

A good view, therefore, simultaneously maximizes $I(V(x); y)$, while preserving discriminative content and $I(V_1(x); V_2(x))$, ensuring sufficient shared structure for alignment. Crucially, the contrastive objective only directly optimizes the latter term. The quality of $I(V(x); y)$ is entirely determined by the view construction strategy, making its design a critical, non-trivial choice.

Many augmentations commonly applied to time series, including jittering, scaling, channel permutation, and temporal masking, fail one or more of these constraints (Figure \ref{fig: perturb}). Jittering introduces additive noise that corrupts shared latent content, reducing $I(V_1; V_2)$. Channel permutation destroys the multivariate dependency structure, which is often class-informative. Masking removes the observed signal, strictly decreasing $I(V; y)$ when the masked content is discriminative. Table~\ref{tab:augmentation_comparison} summarizes these trade-offs.

\begin{table}[t]
\centering
\caption{Comparison of view construction strategies under the label preservation and class informativeness criteria. \checkmark~indicates the property is preserved; \textbf{$\times$} indicates it is violated; $\sim$ indicates partial or conditional preservation.}
\label{tab:augmentation_comparison}
\renewcommand{\arraystretch}{1.2}
\resizebox{\textwidth}{!}{ 
\begin{tabular}{lccc}
\hline
\textbf{Strategy} & \textbf{Preserves} $I(V;\,y)$ & \textbf{Preserves} $I(V_1;\,V_2)$ & \textbf{Domain Assumptions} \\
\hline
Temporal Shift   & \checkmark (maximally) & \checkmark (naturally) & Minimal \\
Jitter           & $\times$               & $\sim$                 & Noise model required \\
Channel Permutation & $\times$            & $\times$               & Domain-specific \\
Masking          & $\sim$                 & \checkmark             & Mask ratio tuning \\
\hline
\end{tabular}
}
\end{table}

Temporal shifting is the only strategy that preserves both information channels without imposing domain-specific assumptions, motivating its use as the basis for our view construction. Furthermore, by minimizing $\mathcal{L}_\text{InfoNCE}$, the encoder maximizes 
a lower bound on $I(V_1;\, V_2) \geq \log N - \mathcal{L}_\text{InfoNCE}$ \cite{cpc}, and since temporal shifting 
preserves $I(V; y)$ without corrupting shared latent content, it provides the 
tightest achievable lower bound among the view construction strategies 
considered.

\subsection{Shift-Based Temporal View Construction}
\label{sec:view_construction}
Given a time series instance $x \in \mathbb{R}^{T \times C}$, we construct 
two views by partitioning the sequence into overlapping sub-sequences at a 
fixed split point. Given overlap ratio $\rho \in (0, 1)$, the view length 
and split indices are fully determined by the sequence length $T$:
\begin{equation}
    L = \frac{T}{2 - \rho}, \qquad L_1 = T - L, \qquad L_2 = L.
\end{equation}
The two views are then defined as: $
    x^{(1)} = x_{:L_2}, \qquad x^{(2)} = x_{L_1:} $,
where $x_{:L_2}$ denotes the prefix up to index $L_2$ and $x_{L_1:}$ denotes 
the suffix from index $L_1$. The overlap region $x_{L_1:L_2}$ is shared 
between both views, providing a common anchor of semantic content, while the 
non-overlapping regions $x_{:L_1}$ and $x_{L_2:}$ introduce sufficient 
divergence to prevent trivial identity matching. The temporal shift 
$\delta = L_1 = \frac{T(1-\rho)}{2-\rho}$ is therefore not a free parameter 
but a consequence of the overlap ratio and sequence length. At the default 
$\rho = 0.5$, this yields $L = \frac{2T}{3}$, reducing the effective sequence 
length processed per view by one third compared to the full sequence, directly accounting for the training speedup reported in Section~\ref{sec:efficiency}. This formulation is entirely deterministic: given a fixed overlap ratio, the split indices are computed without any stochasticity, eliminating the need for augmentation sampling or hyperparameter search over transformation magnitudes.

\subsection{Normalized Temperature-Scaled Cross-Entropy Loss}

Given a batch of $N$ instances $\{x_i\}_{i=1}^N$, we construct two views $\{x_i^{(1)}\}$ and $\{x_i^{(2)}\}$ via the shift-based construction described above. Each view is encoded and projected to obtain normalized embeddings $z_i^{(1)} = f_\theta(x_i^{(1)})$ and $z_i^{(2)} = f_\theta(x_i^{(2)})$. The contrastive objective is the normalized temperature-scaled cross-entropy loss (NT-Xent) \cite{simclr}:

\begin{equation}
    \mathcal{L}_i = -\log \frac{\exp\!\left(\mathrm{sim}(z_i^{(1)}, z_i^{(2)}) / \tau\right)}{\displaystyle\sum_{n=1}^{N} \mathbf{1}_{[n \neq i]}\, \exp\!\left(\mathrm{sim}(z_i^{(1)}, z_n^{(2)}) / \tau\right)},
\end{equation}

where $\mathrm{sim}(u, v) = u^\top v / (\|u\| \|v\|)$ is the cosine similarity and $\tau > 0$ is a temperature hyperparameter. The overall loss is computed symmetrically and averaged over all instances in the batch:

\begin{equation}
    \mathcal{L} = \frac{1}{2N} \sum_{i=1}^{N} \left[\mathcal{L}(z_i^{(1)}, z_i^{(2)}) + \mathcal{L}(z_i^{(2)}, z_i^{(1)})\right].
\end{equation}
All $2(N-1)$ remaining views in the batch serve as negatives for each anchor, requiring no explicit negative sampling or memory bank. Furthermore, the InfoNCE objective provides a lower bound on the mutual information between views~\cite{cpc}:
\begin{equation}
    I(V_1;\, V_2) \geq \log N - \mathcal{L}_\text{InfoNCE},
\end{equation}
such that minimizing $\mathcal{L}$ directly maximizes this bound. Since temporal shifting preserves $I(V;\, y)$ without corrupting shared latent content, it provides the tightest achievable lower bound among the considered view construction strategies. This design, combined with the deterministic view construction, makes the training pipeline fully end-to-end and free of additional hyperparameters beyond the temperature $\tau$ and the overlap ratio $\rho$, which we fix at $\rho = 0.5$ and $\tau = 0.1$ across all experiments.

\begin{figure}[ht]
\includegraphics[width=\textwidth]{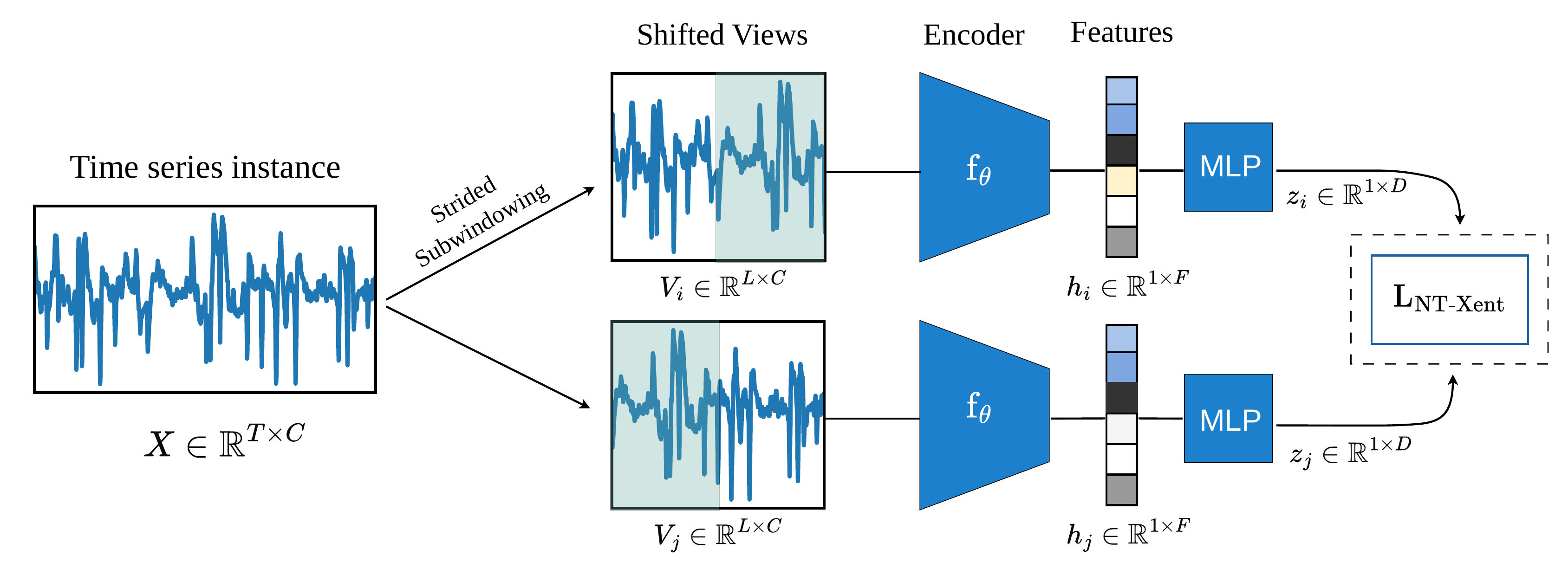}
\caption{\textbf{Overview of ShiFT.} A time series instance $X \in 
\mathbb{R}^{T \times C}$ is partitioned into two overlapping shifted views 
$V_i, V_j \in \mathbb{R}^{L \times C}$ via strided subwindowing. Each view 
is independently encoded by a shared encoder $f_\theta$ into a representation 
$h \in \mathbb{R}^{1 \times F}$, then projected by an MLP head into 
$z \in \mathbb{R}^{1 \times D}$. The NT-Xent loss $\mathcal{L}_\text{NT-Xent}$ 
is computed over the projected representations. The MLP head is discarded at 
evaluation time.}
\label{fig:ShiFT_diagram}
\end{figure}

\section{Experiments}

\subsection{Evaluation Protocol}

\textbf{Datasets.}
We evaluate our method on six publicly available large-scale real-world time series datasets: PAMAP2 ~\cite{pamap2}, WISDM2 ~\cite{wisdm2}, HARTH ~\cite{harth},
SLEEP ~\cite{sleepeeg}, ECG ~\cite{ecg}, and SKODA ~\cite{skoda},
spanning heterogeneous temporal dynamics from $T = 40$ to $T = 3000$, and input channels ranging from $C = 1$ to $C = 52$ across activity recognition, physiological monitoring, and industrial sensing. To further assess generalization across diverse temporal domains and 
dataset scales, we additionally evaluate on the univariate UCR~\cite{ucr} and multivariate UEA~\cite{uea}
archives. Full dataset statistics are provided in the Appendix. To evaluate the
quality of learned representations, we follow the widely adopted linear evaluation
protocol~\cite{simclr,ts2vec,catt}, where a linear classifier is trained on top of the frozen
encoder and test accuracy serves as a proxy for representation quality. Beyond linear probing,
we evaluate under low-label regimes, $k$NN classification, and clustering, and provide
qualitative analysis via t-SNE visualizations of the learned embedding space.

\textbf{Baselines.}
We compare our method against the following self-supervised representation learning
baselines: SimCLR~\cite{simclr}, TS2Vec~\cite{ts2vec}, InfoTS~\cite{infots},
and SimMTM~\cite{simmtm}. SimCLR serves as a general-purpose contrastive baseline
with augmentation-based view construction, while TS2Vec, InfoTS, and SimMTM represent
the current state of the art in self-supervised time series representation learning.
All baselines are reproduced using their official implementations under a unified
evaluation protocol to ensure fair comparison.

\textbf{Default Settings.}
All methods share the same InceptionTime
backbone~\cite{inceptiontime} and a unified data-loading pipeline with aligned
hyperparameters wherever applicable. For our method, we append a 2-layer MLP
projection head mapping representations from $f=256$ to a $d = 32$ dimensional embedding
space, which is discarded at evaluation time (Figure \ref{fig:ShiFT_diagram}). As the loss, we use
NT-Xent~\cite{simclr}, optimized using AdamW with a learning rate of
$1 \times 10^{-3}$, weight decay $1 \times 10^{-4}$, and momentum parameters
$\beta_1 = 0.9$, $\beta_2 = 0.99$. We employ a cosine learning rate schedule
with linear warm-up over the first $10\%$ of training iterations. For the six
large-scale datasets, models are trained for 1500 iterations with batch size 128;
batch size is reduced to 16 for SLEEP, HARTH, and ECG due to memory
constraints imposed by long sequence lengths. For the \textsc{UCR} and \textsc{UEA}
benchmarks, we follow the training protocol of~\cite{ts2vec}: models are trained
for 600 iterations on datasets with effective size greater than 100{,}000, and for
200 iterations otherwise, where effective size is defined as the product of the
number of samples and the window length, with a batch size of 8 throughout. Our
method uses a fixed overlap ratio of $\rho^{*} = 0.5$ and temperature $\tau = 0.1$
across all experiments without any dataset-specific tuning. All experiments are
conducted on NVIDIA V100 GPUs.

\subsection{Linear Evaluation}

We assess the quality of learned representations by freezing the pretrained backbone
and training a logistic regression classifier on the extracted embeddings. This linear
evaluation protocol provides a direct measure of representation quality, independent
of any fine-tuning. Tables~\ref{tab:linear_results} report results across all six
large-scale datasets. ShiFT achieves the
highest average accuracy of $81.49\%$, outperforming the next best method TS2Vec
($79.72\%$) and SimCLR ($78.84\%$), while requiring less training time than all
competing SSL methods.

\begin{table}[t]
\caption{Linear probing accuracy (\%) and total training time (hours) across the six large-scale datasets, averaged over five independent runs with different random seeds. Best results are in \textbf{bold}, second best are \underline{underlined}.}
\label{tab:linear_results}
\resizebox{\textwidth}{!}{ 
\begin{tabular}{lcccccccc}
\toprule
 & ECG2 & HARTH & PAMAP2 & SKODA & SLEEPM & WISDM2 & Total Time (h) & Avg Acc \\
\midrule

Rand Init. 
& 70.41 $\pm$ 6.09 
& 90.01 $\pm$ 1.20 
& \underline{71.91 $\pm$ 1.93} 
& \textbf{99.12 $\pm$ 0.12} 
& 82.26 $\pm$ 0.64 
& 59.96 $\pm$ 1.41 
& <0.01 
& 78.94 \\

InfoTS 
& 72.36 $\pm$ 1.47 
& 77.99 $\pm$ 6.52 
& 70.65 $\pm$ 2.41 
& 98.98 $\pm$ 0.12 
& 82.13 $\pm$ 0.36 
& 60.90 $\pm$ 2.82 
& 2.91 
& 77.17 \\

SimCLR 
& 74.73 $\pm$ 2.97 
& 83.02 $\pm$ 7.05 
& 71.19 $\pm$ 0.35 
& 96.68 $\pm$ 0.27 
& 85.15 $\pm$ 0.19 
& \underline{62.29 $\pm$ 3.08} 
& 3.11 
& 78.84 \\

SimMTM 
& \underline{76.97 $\pm$ 2.44} 
& 89.99 $\pm$ 2.00 
& 70.15 $\pm$ 0.50 
& 98.19 $\pm$ 0.10 
& \textbf{85.22 $\pm$ 0.21} 
& 60.64 $\pm$ 3.66 
& 10.23 
& \underline{80.20 } \\

TS2Vec 
& 74.61 $\pm$ 8.76 
& \textbf{90.49 $\pm$ 2.67} 
& 70.74 $\pm$ 0.80 
& 98.93 $\pm$ 0.16 
& 84.81 $\pm$ 0.21 
& 59.95 $\pm$ 2.20 
& 3.02 
& 79.92 \\

\midrule

ShiFT (ours) 
& \textbf{77.89 $\pm$ 10.68} 
& \underline{90.34 $\pm$ 3.87} 
& \textbf{72.74 $\pm$ 2.67} 
& \underline{99.08 $\pm$ 0.12} 
& \underline{85.22 $\pm$ 0.30} 
& \textbf{63.68 $\pm$ 1.01} 
& 2.35 
& \textbf{81.49 } \\

\bottomrule
\end{tabular}
}
\end{table}

\textbf{UCR and UEA benchmarks.}
To further assess generalization to diverse temporal characteristics,
we evaluate on the \textsc{UCR} and \textsc{UEA} archives. Table~\ref{tab:ucr_uea} reports
average accuracy and average accuracy rank across all datasets in each benchmark. ShiFT
achieves the best average accuracy on both \textsc{UCR} ($80.99\%$) and \textsc{UEA}
($70.26\%$), along with the best average rank in both cases, demonstrating consistent
generalization across univariate and multivariate settings. ShiFT also achieves this
with substantially lower total training time than all competing SSL methods on both
benchmarks.

\begin{table}[ht]
\centering
\caption{Average accuracy (\%), average accuracy rank, and total training time (h)
on the \textsc{UCR} and \textsc{UEA} benchmarks. Lower rank is better.
\textbf{Bold} denotes best result per column.}
\label{tab:ucr_uea}
\renewcommand{\arraystretch}{1.2}

\resizebox{0.80\textwidth}{!}{
\begin{tabular}{l
c
c@{\hspace{10pt}}
c
@{\hspace{16pt}}
c
c@{\hspace{10pt}}
c}

\toprule
& \multicolumn{3}{c}{\textbf{124 UCR Datasets}} 
& \multicolumn{3}{c}{\textbf{28 UEA Datasets}} \\

\cmidrule(lr){2-4}
\cmidrule(lr){5-7}

Method 
& Avg. Acc. (\%) 
& Avg. Rank 
& Time (h) 
& Avg. Acc. (\%) 
& Avg. Rank 
& Time (h) \\

\midrule

Rand Init.
& 71.64 & 4.71 & <0.01
& 67.28 & 3.79 & <0.01 \\

SimCLR  
& 78.76 & 3.31 & 0.99  
& 68.99 & 3.25 & 0.28 \\

TS2Vec  
& 79.05 & 3.00 & 1.71  
& 69.81 & 3.57 & 0.38 \\

InfoTS  
& 71.93 & 4.83 & 1.46  
& 66.00 & 4.14 & 0.35 \\

SimMTM 
& 79.23 & 2.79 & 1.61  
& 69.58 & 3.36 & 1.38 \\

\midrule

ShiFT (Ours)           
& \textbf{80.99} & \textbf{2.37} & \textbf{0.88}
& \textbf{70.26} & \textbf{2.89} & \textbf{0.19} \\

\bottomrule
\end{tabular}
}

\end{table}

\subsection{\textit{k}NN Evaluation and Clustering}

\textbf{\textit{k}NN evaluation.}
Linear probing measures representation quality through the lens of a trained
classifier, which can mask weaknesses in the embedding space. To complement this,
we evaluate representations directly using a parameter-free $1$-nearest neighbour
($1$NN) classifier, which assigns each test sample the label of its closest training
neighbour under Euclidean distance. No additional training is performed as the frozen
encoder embeddings are used directly. To probe label efficiency, the reference set
consists of a balanced $1\%$ sample of labeled training instances, with all embeddings
standardized prior to retrieval. Results are averaged over five random seeds.

Table~\ref{tab:knn} reports results across the six large-scale datasets. ShiFT
achieves the best average accuracy of $67.17\%$, with consistent improvements over
SimCLR ($63.76\%$), TS2Vec ($64.39\%$), InfoTS ($64.34\%$), and SimMTM ($61.19\%$).
The fact that these gains hold under a non-parametric protocol, where no classifier
can compensate for poor embedding geometry, provides strong evidence that ShiFT
learns well-structured representations in which class membership is reflected in
local neighbourhood structure. Notably, removing temperature scaling ($\tau = 1.0$) 
further improves 1NN accuracy to $70.68\%$, suggesting that softer contrastive 
distributions better preserve local neighbourhood geometry; we nevertheless fix 
$\tau = 0.1$ across all experiments for consistency with the linear probing setting.

\begin{table}[t]
\caption{$1$NN classification accuracy (\%) across the six large-scale datasets, averaged over five independent runs with different random seeds. Best results are in \textbf{bold}, second best are \underline{underlined}. \light{ShiFT ($\tau$=1.0)} is shown in grey for reference, illustrating the effect of temperature scaling on neighbourhood geometry without altering the default configuration.}
\label{tab:knn}
\resizebox{\textwidth}{!}{
\begin{tabular}{lccccccc}
\toprule
 & ECG2 & HARTH & PAMAP2 & SKODA & SLEEPM & WISDM2 & Avg Acc \\
\midrule
Rand Init. & \underline{57.47 $\pm$ 10.08} & \underline{72.72 $\pm$ 7.19} & 55.78 $\pm$ 0.81 & 95.30 $\pm$ 0.44 & 57.62 $\pm$ 1.07 & 51.08 $\pm$ 1.49 & 65.00  \\
InfoTS & 56.70 $\pm$ 3.93 & 70.91 $\pm$ 7.60 & \textbf{59.11 $\pm$ 2.85} & \underline{95.33 $\pm$ 0.99} & 58.68 $\pm$ 0.83 & 52.89 $\pm$ 1.59 & 65.61  \\
SimCLR & 56.47 $\pm$ 3.62 & 66.50 $\pm$ 6.90 & 56.90 $\pm$ 0.84 & 85.02 $\pm$ 0.49 & \underline{64.90 $\pm$ 1.41} & 52.77 $\pm$ 1.20 & 63.76  \\
SimMTM & 42.69 $\pm$ 4.35 & 55.73 $\pm$ 3.49 & 54.75 $\pm$ 0.68 & 86.72 $\pm$ 0.85 & 64.51 $\pm$ 0.91 & 50.67 $\pm$ 1.89 & 59.18  \\
TS2Vec & 47.12 $\pm$ 12.39 & \textbf{81.13 $\pm$ 1.63} & 57.90 $\pm$ 0.75 & 91.55 $\pm$ 1.31 & 64.13 $\pm$ 1.24 & \underline{53.04 $\pm$ 1.21} & \underline{65.81 } \\
\midrule
ShiFT (Ours) & \textbf{58.83 $\pm$ 8.07} & 69.40 $\pm$ 9.10 & \underline{58.87 $\pm$ 0.95} & \textbf{97.22 $\pm$ 0.29} & \textbf{65.36 $\pm$ 1.48} & \textbf{53.33 $\pm$ 0.77} & \textbf{67.17 } \\

\light{ShiFT (Ours, $\tau$=1.0)} &
\light{63.74 $\pm$ 8.53} &
\light{76.13 $\pm$ 7.21} &
\light{66.79 $\pm$ 4.43} &
\light{98.12 $\pm$ 0.25} &
\light{65.36 $\pm$ 1.48} &
\light{53.97 $\pm$ 1.53} &
\light{70.68} \\
\bottomrule
\end{tabular}
}
\end{table}

\textbf{Clustering.}
We further examine the global organisation of the embedding space by applying
$k$-means clustering to frozen representations extracted from each method, using
the ground-truth number of classes as $k$. Cluster quality is measured using
Normalised Mutual Information (NMI) and Adjusted Rand Index (ARI) across the
six large-scale datasets. Table~\ref{tab:clustering} reports per-dataset results
alongside the average rank across both metrics.

ShiFT achieves the best average rank of $1.92$, winning outright on 
four of the six datasets (SKODA, WISDM2, PAMAP2, SLEEP) and placing second on 
ECG. The gains are most pronounced on SKODA ($\text{NMI}=0.930$, 
$\text{ARI}=0.828$) and WISDM2 ($\text{NMI}=0.213$, $\text{ARI}=0.158$), 
where ShiFT leads all baselines by a clear margin. On HARTH and ECG, 
SimCLR and InfoTS respectively achieve higher absolute NMI and ARI values, 
suggesting that these datasets may benefit from augmentation-based invariances 
beyond temporal shifting. Nevertheless, the consistent advantage in average rank 
across all six datasets confirms that ShiFT produces the most 
globally well-organised embedding space overall. This is consistent with the 
$k$NN findings and reinforces the same conclusion: temporal shift invariance 
as an inductive bias produces features whose structure reflects semantic 
class boundaries both locally and globally in the embedding space.

\begin{table}[t]
\centering
\caption{Clustering performance (NMI / ARI) across the six large-scale datasets, averaged over five independent runs with different random seeds. Best results are in \textbf{bold}, second best are \underline{underlined}. Average rank is computed across both metrics; lower rank is better.}

\resizebox{\textwidth}{!}{%
\begin{tabular}{lcc@{\hspace{10pt}}cc@{\hspace{10pt}}cc@{\hspace{10pt}}cc@{\hspace{10pt}}cc@{\hspace{10pt}}cc@{\hspace{10pt}}c}
\toprule
Method
& \multicolumn{2}{c}{ECG}
& \multicolumn{2}{c}{HARTH}
& \multicolumn{2}{c}{SKODA}
& \multicolumn{2}{c}{WISDM2}
& \multicolumn{2}{c}{PAMAP2}
& \multicolumn{2}{c}{SLEEP}
& Rank \\

\cmidrule(lr){2-3}
\cmidrule(lr){4-5}
\cmidrule(lr){6-7}
\cmidrule(lr){8-9}
\cmidrule(lr){10-11}
\cmidrule(lr){12-13}

 & {\scriptsize NMI} & {\scriptsize ARI}
 & {\scriptsize NMI} & {\scriptsize ARI}
 & {\scriptsize NMI} & {\scriptsize ARI}
 & {\scriptsize NMI} & {\scriptsize ARI}
 & {\scriptsize NMI} & {\scriptsize ARI}
 & {\scriptsize NMI} & {\scriptsize ARI}
 & Avg \\

\midrule

Rand Init.
& 0.264 & \underline{0.220}
& 0.675 & 0.466
& \underline{0.910} & \underline{0.825}
& 0.144 & 0.104
& 0.575 & 0.378
& 0.251 & 0.143
& \underline{3.67} \\

InfoTS
& \textbf{0.284} & \textbf{0.272}
& 0.632 & 0.419
& 0.904 & 0.817
& 0.120 & 0.064
& 0.584 & \textbf{0.395}
& 0.251 & 0.145
& 3.83 \\

SimCLR
& 0.243 & 0.180
& \textbf{0.686} & \textbf{0.499}
& 0.847 & 0.795
& 0.167 & 0.123
& 0.571 & 0.341
& \underline{0.279} & \underline{0.161}
& \underline{3.67} \\

SimMTM
& 0.240 & 0.213
& 0.652 & 0.450
& 0.854 & 0.781
& \underline{0.185} & 0.148
& 0.531 & 0.347
& 0.264 & 0.156
& 4.17 \\

TS2Vec
& 0.195 & 0.150
& \underline{0.686} & \underline{0.498}
& 0.859 & 0.733
& 0.178 & \underline{0.150}
& \underline{0.590} & 0.369
& 0.263 & 0.156
& 3.75 \\

\midrule

ShiFT (ours)
& \underline{0.280} & 0.212
& 0.672 & 0.461
& \textbf{0.930} & \textbf{0.828}
& \textbf{0.213} & \textbf{0.158}
& \textbf{0.613} & \underline{0.379}
& \textbf{0.280} & \textbf{0.170}
& \textbf{1.92} \\

\bottomrule
\end{tabular}%
}
\label{tab:clustering}
\end{table}

\subsection{Computational Efficiency}
\label{sec:efficiency}
Beyond representation quality, a distinguishing property of ShiFT 
is its computational efficiency.
Figure~\ref{fig:radar} reports total training time across the six 
large-scale datasets as well as the \textsc{UCR} and \textsc{UEA} benchmarks. 
ShiFT achieves the lowest training time in all three settings, 
outpacing not only methods with complex sampling and masking strategies such as SimMTM and TS2Vec, but also SimCLR, which shares the same NT-Xent objective.

This last comparison is particularly revealing. Since ShiFT and 
SimCLR differ only in their view construction strategy, the speed advantage 
cannot be attributed to architectural differences or a simpler loss. Instead, 
it arises directly from the split view construction: as shown in 
Section~\ref{sec:view_construction}, at $\rho = 0.5$ each view has effective 
length $L = \frac{2}{3}T$, meaning the encoder processes only $66.7\%$ of 
the original sequence per view. Since the encoder cost scales with the sequence 
length, this yields a proportional reduction in training time, a free 
efficiency gain that requires no approximation or architectural modification. This efficiency gain is not accompanied by any degradation in representation 
quality. Figure~\ref{fig:radar} presents a radar chart comparing all methods 
across six downstream evaluation axes: linear probing (Large, UCR, UEA), 
$1$NN accuracy, and clustering (NMI, ARI), where ranks are inverted such 
that a larger enclosed area indicates consistently lower average rank. 
ShiFT achieves the largest enclosed area across all axes, 
demonstrating that its computational advantage does not come at the cost of 
representational quality.

\section{Analysis}

\subsection{Architecture and Projection Head}

\begin{figure}[t]
\centering
\begin{minipage}{0.44\linewidth}
\centering
\setlength{\tabcolsep}{10pt} 
\captionof{table}{Ablation study of ShiFT on the six large-scale datasets. Linear probing accuracy (\%) and $1$NN accuracy (\%) are reported as averages over five independent runs with different random seeds.}
\resizebox{\linewidth}{!}{
\begin{tabular}{lcc}
\toprule
 & \textbf{Linear} & \textbf{kNN}\\
\midrule

\multicolumn{3}{l}{\textit{Backbone $f_\theta$}} \\

\quad $\rightarrow$ FCN & 79.58 & 68.65 \\
\quad $\rightarrow$ ResNet & 79.12 & 66.90 \\
\quad $\rightarrow$ InceptionTime & 81.49 & 67.17 \\

\midrule

\multicolumn{3}{l}{\textit{Augmentations}} \\

\quad $\rightarrow$ Jitter + Scale & 78.84 & 63.76 \\
\quad $\rightarrow$ Random Shift & 80.50 & 62.62 \\
\quad $\rightarrow$ Jitter + Scale + Random Shift & 79.24 & 64.07 \\
\quad $\rightarrow$ ShiFT (Ours) & \textbf{81.49} & \textbf{67.17} \\

\bottomrule
\label{tab:ablation1}
\end{tabular}
}
\label{ablation-ucr-uea}
\end{minipage}
\hfill
\begin{minipage}{0.55\linewidth}
\centering
\includegraphics[width=\linewidth]{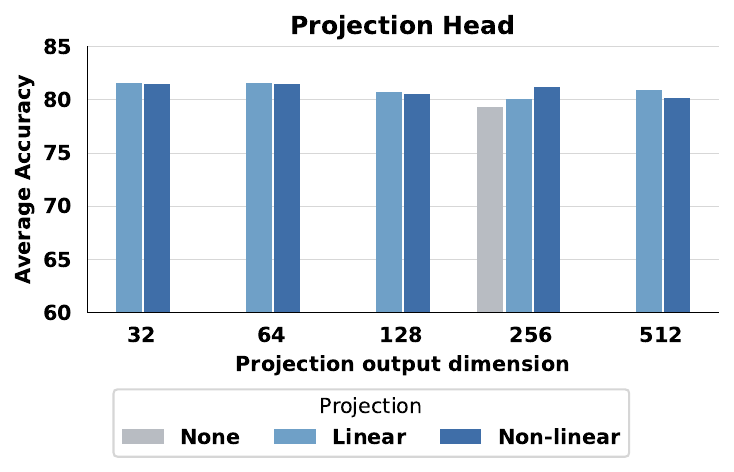}
\caption{Effect of projection head type and output dimension on linear probing 
accuracy (\%), averaged over the six large-scale datasets.}
\label{fig:projection}
\end{minipage}

\end{figure}

We study the effect of encoder architecture and projection head design on the 
quality of learned representations. Table~\ref{tab:ablation1} compares three 
backbone architectures: FCN, ResNet, and InceptionTime, under linear 
evaluation and $1$NN on the six large-scale datasets. InceptionTime achieves the best linear probing accuracy ($81.49\%$), while 
FCN achieves the best $1$NN accuracy ($68.65\%$), suggesting a mild 
trade-off between linear separability and neighbourhood geometry across 
backbone choices. ResNet1D underperforms both FCN and InceptionTime on both 
metrics, indicating that residual connections alone do not confer an advantage 
in the time series SSL setting.

Regarding projection head design, both linear and non-linear projection heads 
outperform the no-projection baseline (79.27\% at $d=256$), confirming that 
the head absorbs contrastive-specific invariances and leaves the encoder 
output enriched with transferable features, consistent with findings in 
visual SSL~\cite{simclr}. However, contrary to the clear margin that 
non-linear heads provide in vision, we find the two are largely equivalent 
in the time series setting, with differences within the noise range across all 
projection dimensions $d \in \{32, 64, 128, 256, 512\}$. This suggests that 
the representational bottleneck in time series contrastive learning lies in 
the view construction strategy rather than the expressive capacity of the 
projection head. Performance is also robust to projection dimensionality, 
with a slight preference for smaller dimensions ($d = 32, 64$), suggesting 
that time series representations occupy a more compact latent structure than 
natural images. The projection head is discarded at evaluation time in all 
experiments.

\subsection{Temporal Shift Invariance as Inductive Bias}

Table~\ref{tab:ablation1} isolates the effect of view construction by comparing ShiFT against 
augmentation-based alternatives on the six large-scale datasets under identical backbone, loss, and training configuration. ShiFT achieves the best performance on both metrics, 
outperforming Jitter + Scale by $2.65$ points on linear probing and $3.41$ 
points on $1$NN. Specifically, Random Shift, which applies circular shifts of random 
magnitude, producing views of identical length $T$ with no controlled overlap 
structure, achieves competitive linear probing ($80.50\%$) but degrades 
$1$NN accuracy to $62.62\%$, the lowest among all configurations. Beyond the 
inconsistent semantic sharing between views, circular rolling introduces 
artificial discontinuities at the wraparound boundary and fails the 
non-triviality criterion: at small shift magnitudes the two views are nearly 
identical, collapsing the contrastive task. ShiFT avoids both 
failure modes by construction, yielding representations that are both linearly separable 
and geometrically well-structured.

\subsection{Batch Size in Time Series Contrastive Learning}

A well-established finding in contrastive learning for vision is that larger batch
sizes improve performance by providing more negative examples per update
\cite{simclr,moco}. We investigate whether this holds for time series by varying
the batch size over $\{32, 64, 128, 256, 512\}$ and evaluating linear probing
accuracy on four datasets: \textsc{HARTH}, \textsc{SKODA}, \textsc{PAMAP2}, and
\textsc{WISDM2}. Figure~\ref{fig:batch_size} reports results for each dataset.

Contrary to observations in CV, we find no consistent benefit from larger 
batch sizes in the time series setting. Performance peaks at a moderate batch 
size of $128$--$256$ on most datasets and degrades at $512$, with the 
exception of SKODA, which increases monotonically across all batch sizes. We attribute the general degradation at large batch sizes to a key 
distributional difference: unlike natural images, time series instances within 
a batch are often drawn from a small number of activity classes with high 
intra-class similarity. As batch size grows, the proportion of false negatives, instances from the same class treated as negatives, increases, 
undermining the contrastive signal. The sharp degradation on HARTH at batch 512 
($84.63\%$ vs $90.33\%$ at batch 128) is particularly illustrative, as HARTH 
contains activities with highly similar motion profiles. This suggests that 
the standard large-batch recipe from vision does not transfer directly to time 
series, and that moderate batch sizes of $128$--$256$ are sufficient and 
preferable in this setting.

\subsection{Temperature and Overlap Ratio}

\begin{figure}[t]
\centering

\begin{minipage}{0.55\linewidth}
\centering
\includegraphics[width=\linewidth]{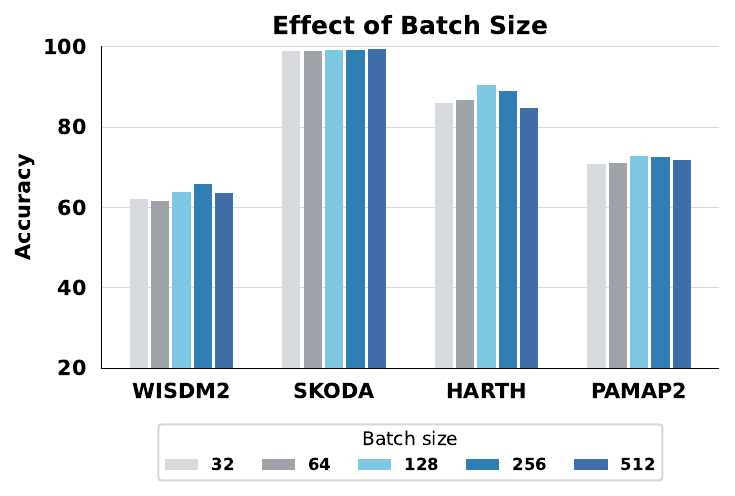}
\caption{Effect of batch size on linear probing accuracy (\%) across four 
representative large-scale datasets.}
\label{fig:batch_size}
\end{minipage}
\hfill
\begin{minipage}{0.40\linewidth}
\centering
\setlength{\tabcolsep}{15pt} 
\captionof{table}{Ablation of temperature $\tau$ and overlap ratio $\rho$ on the six large-scale datasets, reporting linear probing and $1$NN accuracy (\%) averaged over five independent runs.}
\resizebox{\linewidth}{!}{
\begin{tabular}{lcc}
\toprule
 & \textbf{Linear} & \textbf{kNN}\\
\midrule

\multicolumn{3}{l}{\textit{Temperature $\tau$}} \\

\quad $\rightarrow$ 0.1 & 81.49 & 67.17 \\
\quad $\rightarrow$ 0.5 & 81.60 & 67.81 \\
\quad $\rightarrow$ 1.0 & 81.13 & 70.68 \\

\midrule
\multicolumn{3}{l}{\textit{Overlap $\rho$}} \\

\quad $\rightarrow$ 0.25 & 82.18 & 66.87 \\
\quad $\rightarrow$ 0.50 & 81.49 & 67.17 \\
\quad $\rightarrow$ 0.75 & 81.19 & 65.92 \\

\bottomrule
\end{tabular}
}
\label{tab:ablation2}
\end{minipage}

\end{figure}

Table~\ref{tab:ablation2} reports the effect of temperature $\tau$ and overlap 
ratio $\rho$ on linear probing and $1$NN accuracy across the six large-scale 
datasets. Linear probing accuracy is largely insensitive to 
temperature, varying by less than $0.5$ points across $\tau \in \{0.1, 0.5, 
1.0\}$. In contrast, $1$NN accuracy increases substantially with temperature, 
with a notable gain of $2.87$ points between $\tau = 0.1$ ($67.17\%$) and 
$\tau = 1.0$ ($70.68\%$). We attribute this asymmetry to the effect of 
temperature on embedding geometry: low temperatures concentrate the gradient 
signal on the hardest negatives, producing tight, linearly separable clusters 
at the cost of distorting the broader neighbourhood structure; higher 
temperatures distribute the learning signal more uniformly, better preserving 
local geometry and thereby improving nearest-neighbour retrieval. We adopt 
$\tau = 0.1$ as the default, as it achieves competitive linear probing 
accuracy and is consistent with common practice in contrastive 
learning~\cite{simclr}.

Performance decreases monotonically as overlap increases beyond $\rho = 0.25$, with the sharpest drop at $\rho = 0.75$ 
($81.19\%$ linear, $65.92\%$ $1$NN) where views become near-identical and 
the contrastive task degenerates toward trivial similarity matching. The 
results are robust across $\rho \in \{0.25, 0.50\}$, with differences of 
less than $0.7$ points on both metrics, suggesting that the contrastive 
objective is tolerant of moderate variation in overlap provided views remain 
sufficiently distinct. We adopt $\rho = 0.5$ as the default for its balance 
between view diversity and shared temporal content, and note that the only 
consistently harmful configuration is high overlap at $\rho = 0.75$.

\subsection{t-SNE Visualisation of Learned Embeddings}

\begin{figure}[t]
\includegraphics[width=\textwidth]{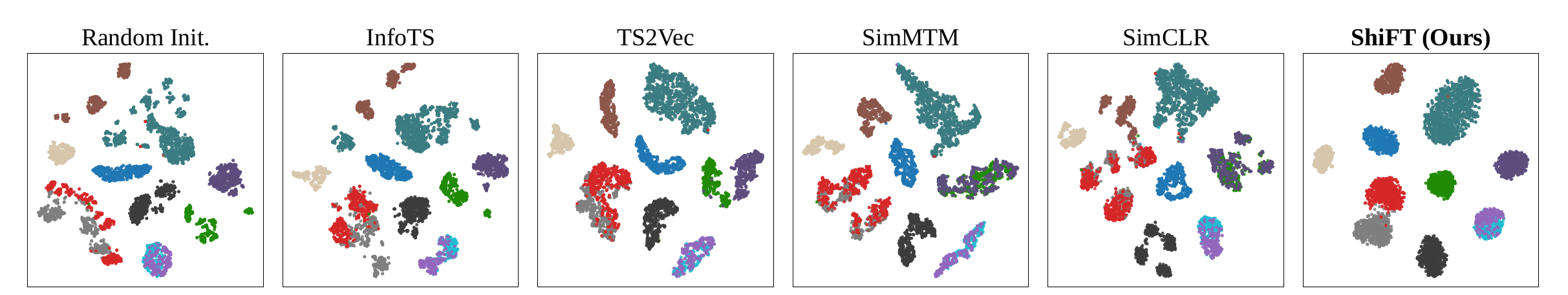}
\caption{t-SNE projections of learned representations on the \textsc{SKODA} 
test set for all methods. Best viewed in 
colour.}
\label{fig:tsne_plots}
\end{figure}

Figure~\ref{fig:tsne_plots} presents t-SNE projections of the learned
representations on the \textsc{SKODA} test set. ShiFT consistently produces compact,
well-separated clusters across all activity classes, while competing methods exhibit
noticeable overlap or fragmentation between classes. Notably, ShiFT is the only method
that achieves clean separation of the most challenging \texttt{\textcolor{red}{red}} and \texttt{\textcolor{gray}{gray}} classes, which remain partially entangled
in the representations of all baselines. The uniform compactness and clear inter-class
margins observed across ShiFT's clusters indicate that deterministic shift-based view
construction induces a highly structured embedding space, capturing fine-grained
temporal distinctions that are not recovered by augmentation-based or masking-based
approaches.

\section{Conclusion}
We presented ShiFT, a self-supervised contrastive learning framework 
for time series classification that replaces hand-crafted augmentations and 
complex view construction strategies with a single principled design choice: 
deterministic shifted-split views with a fixed overlap. We demonstrated that 
temporal shift invariance alone is a sufficient and effective inductive bias 
for learning strong transferable representations, achieving state-of-the-art 
performance across linear probing, $k$NN, and clustering evaluations on six 
large-scale benchmarks and the full \textsc{UCR} and \textsc{UEA} archives, 
while achieving the lowest training time across all settings. Beyond the empirical results, our analysis surfaces an 
important finding: core assumptions from visual contrastive learning, 
particularly the benefit of large batch sizes, do not transfer to time 
series, where high intra-class similarity inflates the proportion of false 
negatives. Together, these 
results suggest that for time series, the right inductive bias applied simply 
and deterministically outperforms elaborate engineering. We hope ShiFT 
serves as both a strong practical baseline and a conceptual argument for 
principled simplicity in temporal representation learning.



%
%
%
\bibliographystyle{splncs04}
\bibliography{mybibliography}
%
\end{document}